\documentclass[12pt,a4paper]{article}

\usepackage{natbib}
\setcitestyle{round}

\usepackage[english]{babel}
\usepackage[utf8]{inputenc}
\usepackage[T1]{fontenc}

\usepackage{graphicx}     % Comando \includegraphics
\usepackage[table]{xcolor} % Adicione isso no preâmbulo
\usepackage{indentfirst}  % Indenta o primeiro parágrafo de cada seção
\usepackage{url}          % Comandos \url e \href
\usepackage[top=2cm, bottom=2cm, left=2cm, right=2cm]{geometry} % Define as margens do documento
\usepackage{multirow}     % Permite criar tabelas com uma célula ocupando várias linhas
\usepackage{amssymb}      % Símbolos matemáticos
\usepackage{amsmath}      % Ambientes para escrever fórmulas, \begin{align} por exemplo.
\usepackage{caption}      % Para definir o estilo das legendas de figuras e tabelas.
\usepackage{setspace}     % Para definir espaçamento entre linhas. (\onehalfspacing, \singlespacing, \doublespacing)
\usepackage{breakcites}   % Para permitir quebra de linha no meio de citações.
\usepackage{times}        % Fonte Times New Roman
\usepackage{lipsum}       % Para gerar texto temporário. Exemplo: \lipsum \lipsum[1] \lipsum[4-5].
\usepackage{inconsolata}  % Fonte boa para códigos e URLs. Use \texttt{}
\usepackage{hyperref}     % Faz os links ficarem azuis e clicáveis. Facilita a navegação pelo PDF.

\title{Effects of label noise on the classification of outlier observations}

\author{
	Matheus Vinícius Barreto de Farias\thanks{Instituto de Ciências Matemáticas e de Computação, Bacharelado em Estatística e Ciência de Dados, Universidade de São Paulo, São Carlos, SP, Brasil. \texttt{matheusbarreto@usp.br}} 
	\and
	Mário de Castro\thanks{Instituto de Ciências Matemáticas e de Computação, Departamento de Matemática Aplicada e Estatística, Universidade de São Paulo, São Carlos, SP, Brasil.}
}

\begin{document}
	\maketitle

	\begin{abstract}
	This study investigates the impact of adding noise to the training set classes in classification tasks using the BCOPS algorithm (\textit{Balanced and Conformal Optimized Prediction Sets}), proposed by \citet{guan2022}. The BCOPS algorithm is an application of conformal prediction combined with a machine learning method to construct prediction sets such that the probability of the true class being included in the prediction set for a test observation meets a specified coverage guarantee. An observation is considered an outlier if its true class is not present in the training set. The study employs both synthetic and real datasets and conducts experiments to evaluate the prediction abstention rate for outlier observations and the model’s robustness in this previously untested scenario. The results indicate that the addition of noise, even in small amounts, can have a significant effect on model performance.
	
	\end{abstract}

	\section{Introduction}
	
	The following study presents results obtained from experiments in which, before training a classification model, we added noise to the labels of the training set, so that the information contained in this set is not entirely correct. In fact, most datasets encountered in practical situations contain some degree of noise, which highlights the importance of this type of study for new techniques before implementing them in real-world applications.
	
	In this case, we are interested in measuring the impact of noise addition on BCOPS \citep{guan2022}, a algorithm based on conformal prediction \citep{vovk2005} which, when combined with other machine learning methods, allows the construction of prediction sets for the test set observations in classification tasks. Prediction sets are sets that contain the possible values (for regression tasks) or possible classes (for classification tasks) for new observations. These sets are constructed so that the probability of the true value or class being contained within them meets a coverage guarantee.
	
	In the work developed by \citet{guan2022}, the possibility of using these prediction sets to detect outlier observations -- meaning, observations whose true class was not present during training -- is emphasized. Thus, we aim to measure both the classification coverage and the abstention rate on outlier observations of the BCOPS algorithm under the addition of noise, considering some of the datasets and machine learning algorithms used by \citet{guan2022}.
	
	This paper is organized as follows: In Section~\ref{sec:conformal}, we provide a brief review of the fundamental concepts necessary for understanding the algorithm; in Section~\ref{sec:ruido}, we present the method used for adding noise; in Section~\ref{sec:resultados}, we show the results obtained from the conducted experiments and raise related discussions; finally, Section~\ref{sec:conclusao} summarizes and discusses the work carried out throughout this study.
	
	The implementations of the experiments in the R language \citep{rlang} can be found at \url{https://github.com/mvbdf/bcops-exemplos}.

	\section{Conformal Prediction and the BCOPS Algorithm}
	\label{sec:conformal}
	
	The idea behind conformal prediction, first introduced by \citet{gammerman98} and \citet{vovk2005}, arises from the need to add a layer of reliability to the predictions made by standard predictive methods, providing a certain guarantee that the true value (for regression problems) or true class (for classification problems) of an observation will be included in a set of possible outcomes.
	
	Imagine we have a population divided into different classes, and we wish to predict the class of a new observation—let $x$ denote the feature vector of this observation—based on the characteristics of other samples with known classes. A common approach would be to classify this new observation as belonging to the most probable class using statistical learning methods, such as clustering based on Euclidean distance \citep[for details on conventional machine learning methods, see][]{hastie2009}. In the conformal prediction paradigm, however, our goal is to construct a set, denoted $C(x)$, containing the possible classes to which $x$ may belong, such that the probability of the true class $y$ being included in $C(x)$ is greater than or equal to a predefined coverage guarantee $1-\alpha$, typically $95\%$.
	
	The construction of these sets is carried out through a conformity score function, which measures how similar a new observation is to the observations of each class present in the training set. Thus, given a new observation $x$, for each class in the training set ($\left\{1, 2, \dots, K\right\}$), we obtain a conformity score that determines whether that class should be included in $C(x)$—the higher the conformity score for a given class, the closer $x$ is to the observations of that class. Among the measures used to quantify conformity, one may use the Euclidean distance itself or, in the case of BCOPS, a machine learning algorithm capable of separating the classes. Details on the different ways to construct the conformity score function can be found in \citet{lei2015} and \citet{guan2022}.
	
	Note that if the conformity score of an observation $x$ is low for all classes present in the training set, no class will be included in $C(x)$, and $x$ may be considered an outlier—thus, the algorithm is said to abstain from making a prediction. The abstention rate on outlier observations is one of the performance metrics used to compare results, as the test set includes observations that do not belong to any of the training classes.
	
	Further information on the conformal prediction framework can be found in \citet{lei2013} and \citet{lei2018}.

	\section{Perturbation by Noise}
	\label{sec:ruido}
	
	A common problem in annotated datasets is the presence of errors in the labels of observations. To examine the behavior of the BCOPS algorithm in this scenario, we refer to the work of \citet{einbinder2022}, which presents empirical results supporting the robustness of conformal prediction under the addition of noise to the labels of the training set. Our concern here is not only to verify compliance with coverage guarantees but also to measure the behavior of the abstention rate in outlier observations, a characteristic that has not yet been assessed.
	
	Following Section~2.2 of \citet{einbinder2022}, we define a corruption function
	
	\begin{equation}
		\label{eq:corrupcao}
		g\left(y\right) = \left\{ \begin{array}{cl}
			y & , \ \text{with probability}\ 1 - \phi, \\
			Y' & , \ \text{otherwise,}
		\end{array} \right.
	\end{equation}
	
	\noindent which changes the label $y$ of each observation in the training set with probability $\phi$, sampling $Y'$ from the set of $K$ classes present in the training data, $\left\{1, 2, \dots, K\right\}$, according to a uniform distribution. This means that after applying the corruption function to a dataset, we will have approximately $100 \cdot \phi \%$ noise in the training labels.
	
	It is worth noting that our analysis focuses on the simplest case, where an observation is mislabeled in a purely random manner. Other noise patterns in labels can be found in specific applications, some of which are described in \citet{einbinder2022}.

	\section{Results}
	\label{sec:resultados}
	
	We take the corruption function defined in \ref{eq:corrupcao} and, as we increase the noise level, we examine the evolution of the coverage rate for each class, defined as
	
	\begin{equation}
		\label{eq:cobertura}
		\frac{1}{n_k}\sum_{i = 1}^{n_k} \mathbb{I}_{C\left(x_{ik}\right)}\left(y_{ik}\right), \quad k = 1, \dots, K,
	\end{equation}
	
	\noindent where $n_k$ is the number of observations of class $k$ in the test set, $\left(x_{ik}, y_{ik}\right)$ is the $i$-th observation belonging to class $k$ in the test set, $C\left(x_{ik}\right)$ is the prediction set for that observation, and $\mathbb{I}_A\left(a\right)$ is the indicator function:
	\begin{equation}
		\mathbb{I}_A\left(a\right) = \left\{ \begin{array}{cl}
			1 & , \ \text{if}\ a \in A, \\
			0 & , \ \text{if}\ a \not\in A.
		\end{array} \right.
	\end{equation}
	
	This value indicates the proportion of observations of class $k$ for which $k \in C\left(x\right)$, and it should be close to the coverage guarantee. For examples with more than two classes in the training set, we take the mean coverage guarantee across noise levels to more simply represent the evolution of the algorithm’s coverage.
	
	We also measure the evolution of the abstention rate for outlier observations, defined as
	
	\begin{equation}
		\label{eq:deteccao}
		\frac{1}{n_a}\sum_{i \in A} \mathbb{I}_{\left\{\emptyset\right\}}\left(C\left(x_{ia}\right)\right),
	\end{equation}
	
	\noindent where $x_{ia}$ is the $i$-th outlier observation in the test set, $n_a$ is the number of observations whose classes are not present in the training set, and $A$ is the set of indices of these observations in the test set. This value indicates how often the algorithm abstained from making a prediction for an observation whose class was not seen during training, and the closer it is to $1$, the better the model’s ability to detect outlier observations.
	
	The following sections present the datasets used and the results obtained with the BCOPS method. In all cases, we used the Random Forest algorithm as the auxiliary learner \citep[\emph{see}][]{ranger2017} and set the coverage guarantee to $95\%$. The repository available at \url{https://github.com/mvbdf/bcops-exemplos} also contains code for the same examples using the ElasticNet algorithm as the auxiliary learner \citep[\emph{see}][]{glmnet}.

	\subsection{Synthetic Data 1}
	\label{ex1}
	
	We generated pairs of observations $\left(x, y\right)$ belonging to two different classes, where $x \in \mathbb{R}^{10}$ is such that
	
	\begin{equation}
		x_1 \sim
		\left\{ \begin{array}{cl}
			N\left(0,1\right) & ,\ \text{if}\ y = 1,\\
			N\left(3, 0.5\right) & ,\ \text{if}\ y = 2.
		\end{array} \right.
		\ ,\ x_j \sim N\left(0, 1\right),\ j = 2, ..., 10,
	\end{equation}
	
	In the test set, we also include a class of outlier observations defined as
	
	\begin{equation}
		x_2 \sim N\left(3, 1\right) \quad \text{and} \quad x_j \sim N\left(0, 1\right),\ j \neq 2.
	\end{equation}
	
	For the training set, we generated 1000 observations, half belonging to class 1 and the other half to class 2. For the test set, we generated 1500 observations, one third of which come from the outlier distribution, while the remaining observations are equally divided between classes 1 and 2.
	
\begin{figure}[h]
		\centering
		\includegraphics[width=0.7\textwidth]{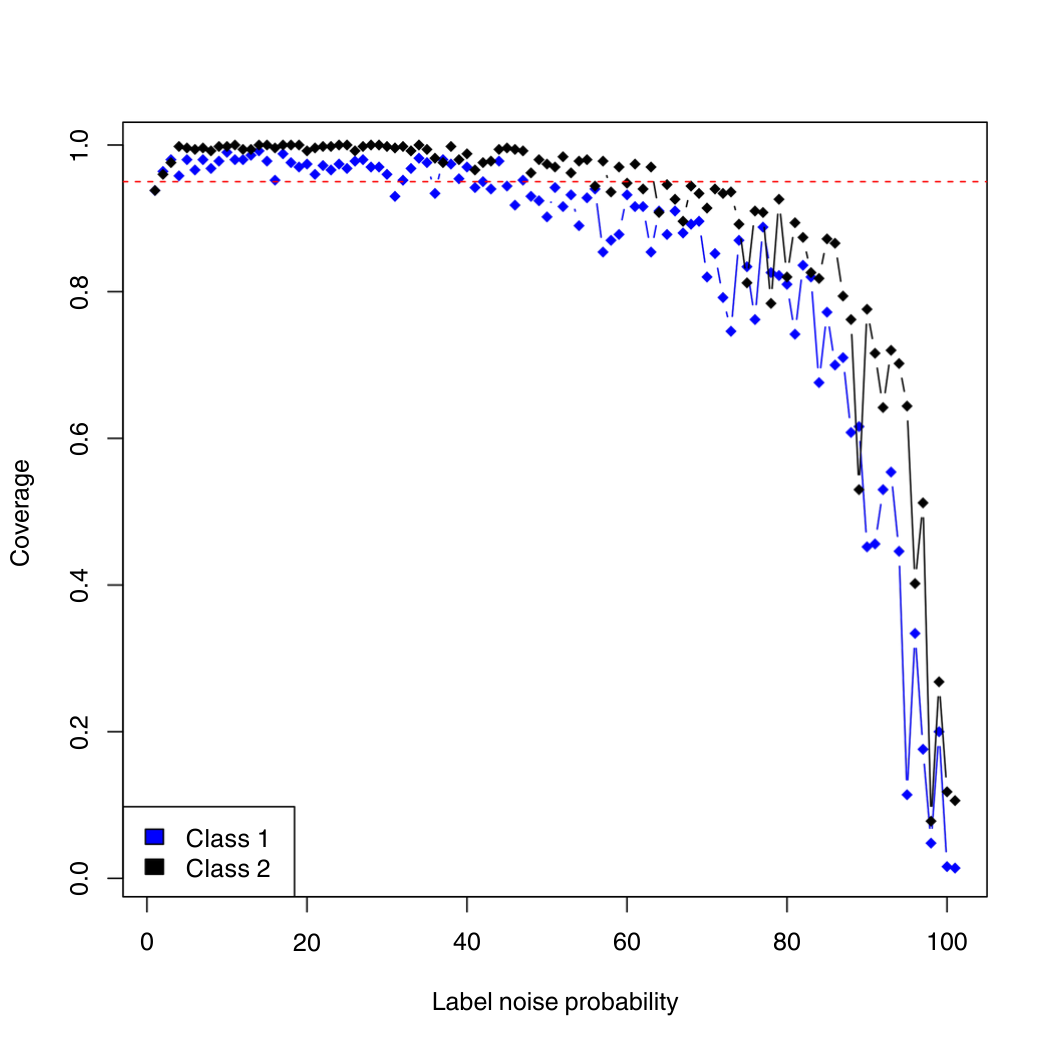}
		\caption{Evolution of the coverage rate by class according to the level of noise in the training set for the dataset generated in Example~\ref{ex1}.}
		\label{img:ex1-rf-cobertura}
	\end{figure}

\begin{figure}[h]
		\centering
		\includegraphics[width=0.7\textwidth]{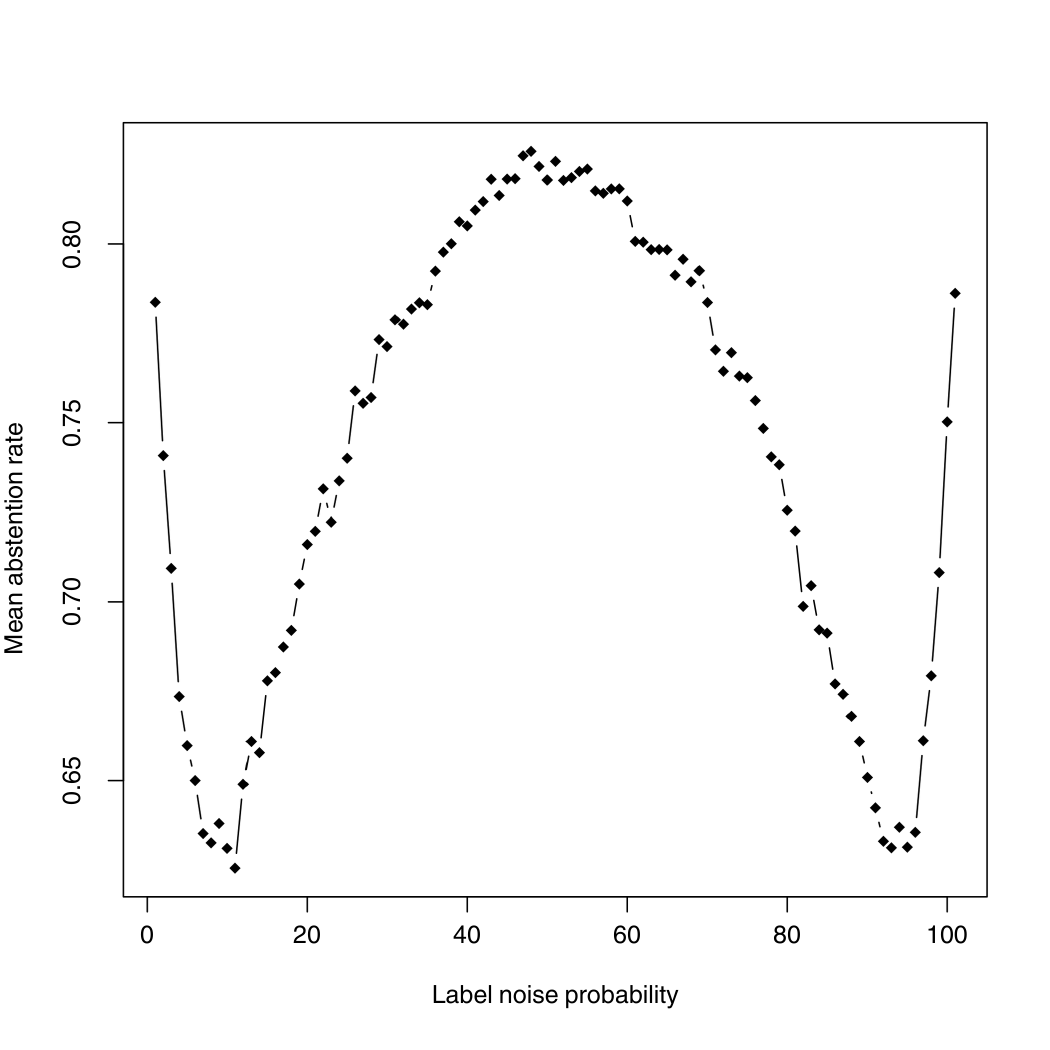}
		\caption{Evolution of the abstention rate for outlier observations according to the level of noise in the training set for the dataset generated in Example~\ref{ex1}.}
		\label{img:ex1-rf-abstencao}
	\end{figure}

	Figure~\ref{img:ex1-rf-cobertura} shows the coverage rates by class according to the level of noise present in the training labels. Figure~\ref{img:ex1-rf-abstencao} presents the evolution of the average abstention rate for outlier observations as the level of noise in the training labels increases. This average was calculated by repeating the label corruption and model training simulation 100 times to produce a more representative depiction of the algorithm’s behavior.
	
	It is also worth noting that in Figure~\ref{img:ex1-rf-abstencao}, the symmetric behavior of the curve is expected and can be explained by the fact that we have two classes whose labels invert as the noise level increases. Consequently, the abstention of the BCOPS algorithm for an outlier observation is affected equivalently for complementary noise levels.
	
	\subsection{Synthetic Data 2}
	\label{ex2}
	
	We generated pairs of observations $\left(x, y\right)$ belonging to ten different classes, where $x \in \mathbb{R}^{10}$ is such that
	
	\begin{equation}
		x_i \sim
		\left\{ \begin{array}{cl}
			N\left(0,1\right) & ,\ \text{if}\ i \neq y,\\
			N\left(3, 0.5\right) & ,\ \text{if}\ i = y.
		\end{array} \right.
		\ ,\ i = 1, 2, ..., 10,
	\end{equation}
	
	\noindent where $y$ is the class of the generated observation. In the test set, we also include a class of outlier observations defined as
	
	\begin{equation}
		x_i \sim N\left(3, 2\right),\ i = 1, 2, ..., 10.
	\end{equation}
	
	In this setup, each class is separable from the others along one of the dimensions, and the outlier class follows the same distribution across all dimensions. Note that this example represents a problem where multiple classes are presented during training and only one class arises from the outlier distribution. For training, we generated $500$ observations for each of the $10$ classes. For testing, in addition to the $500$ observations from each class, we generated another $500$ observations belonging to the outlier class.
	
\begin{figure}[h]
		\centering
		\includegraphics[width=0.7\textwidth]{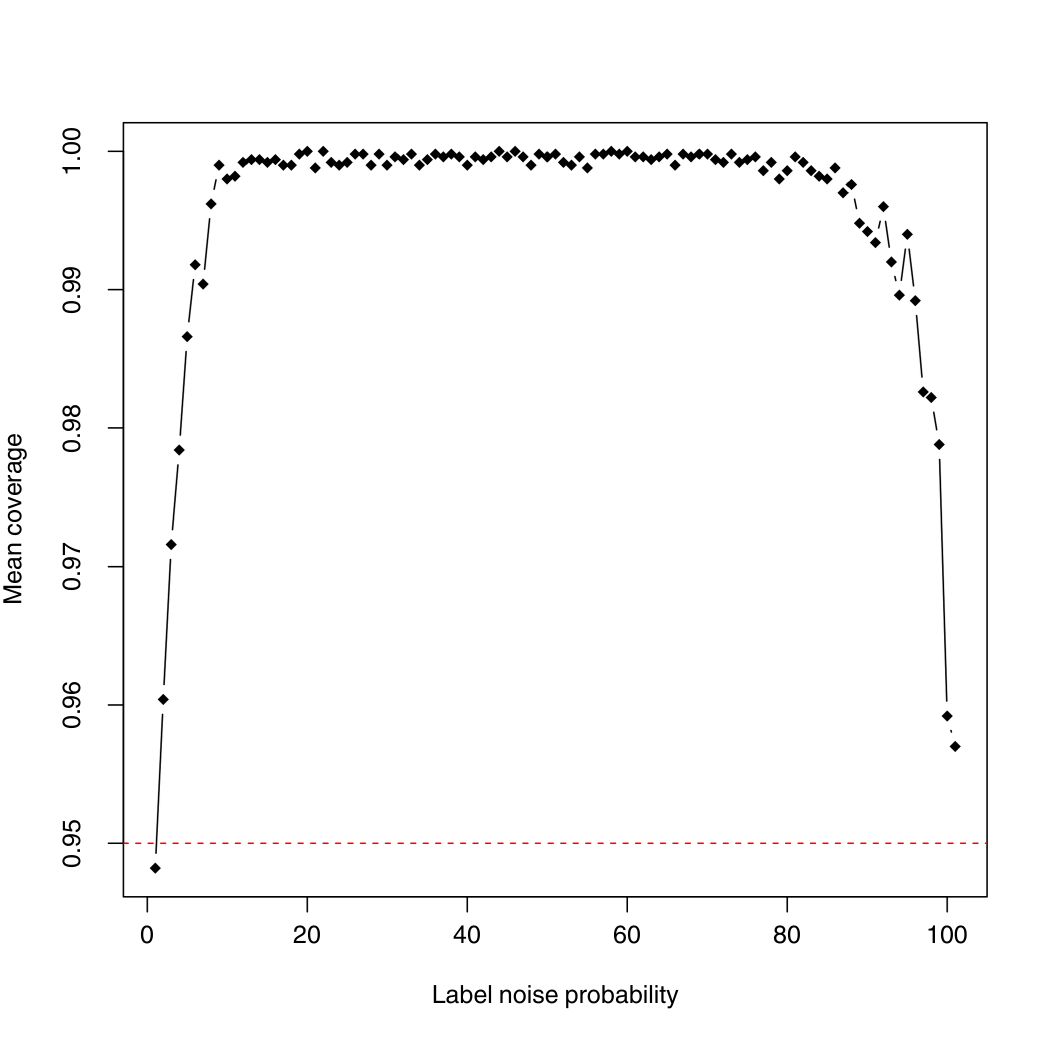}
		\caption{Evolution of the average coverage rate according to the level of noise in the training set for the dataset generated in Example~\ref{ex2}.}
		\label{img:ex2-rf-cobertura}
	\end{figure}

\begin{figure}[h]
		\centering
		\includegraphics[width=0.7\textwidth]{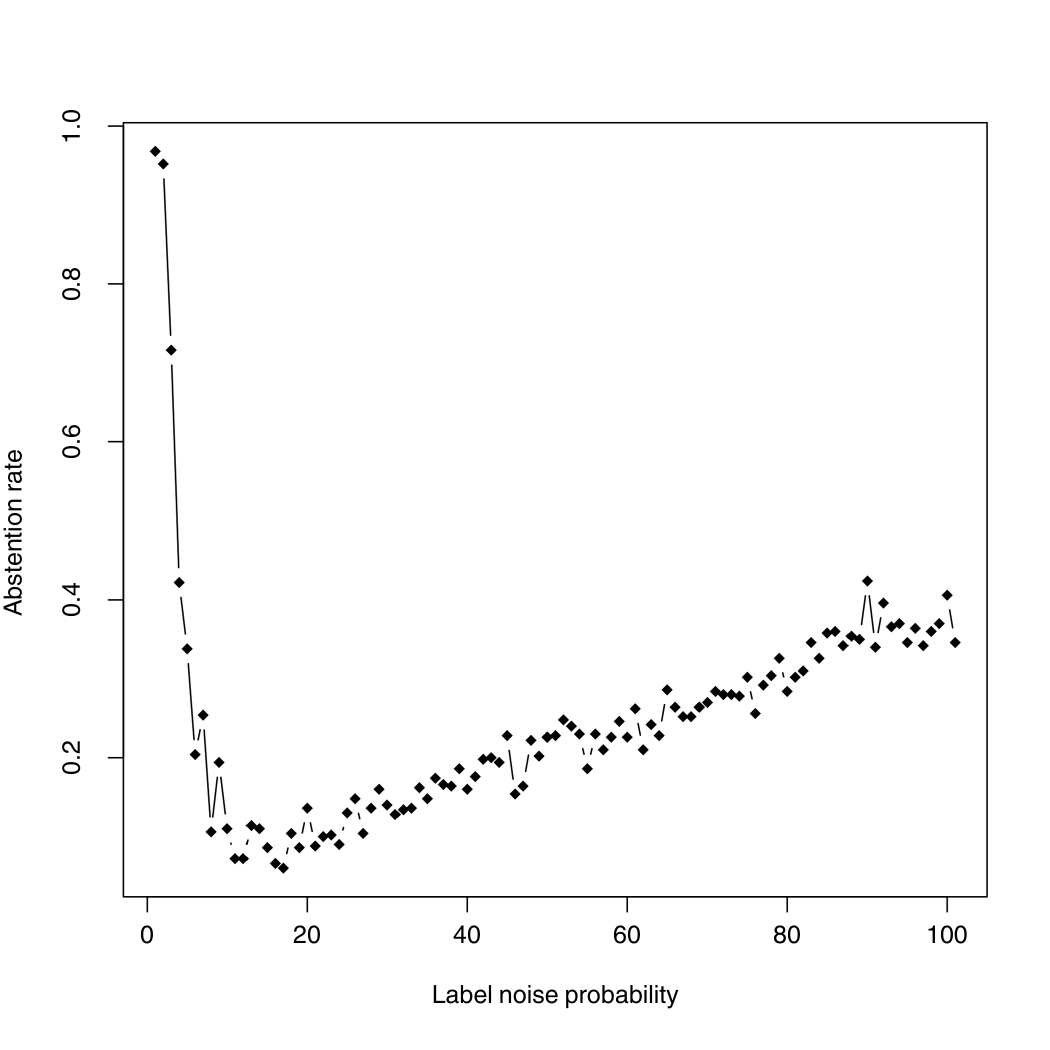}
		\caption{Evolution of the abstention rate for outlier observations according to the level of noise in the training set for the dataset generated in Example~\ref{ex2}.}
		\label{img:ex2-rf-abstencao}
	\end{figure}

	Figure~\ref{img:ex2-rf-cobertura} shows the mean coverage rates per class according to the noise level in the training labels. Figure~\ref{img:ex2-rf-abstencao} presents the evolution of the abstention rate for outlier observations as the noise level in the training labels increases.
	
	\subsection{MNIST}
	\label{exmnist}
	
	The MNIST dataset \citep[see][]{mnist} contains $28\times28$ pixel images of the handwritten digits from $0$ to $9$.
	
\begin{figure}[h]
		\centering
		\includegraphics[width=0.7\textwidth]{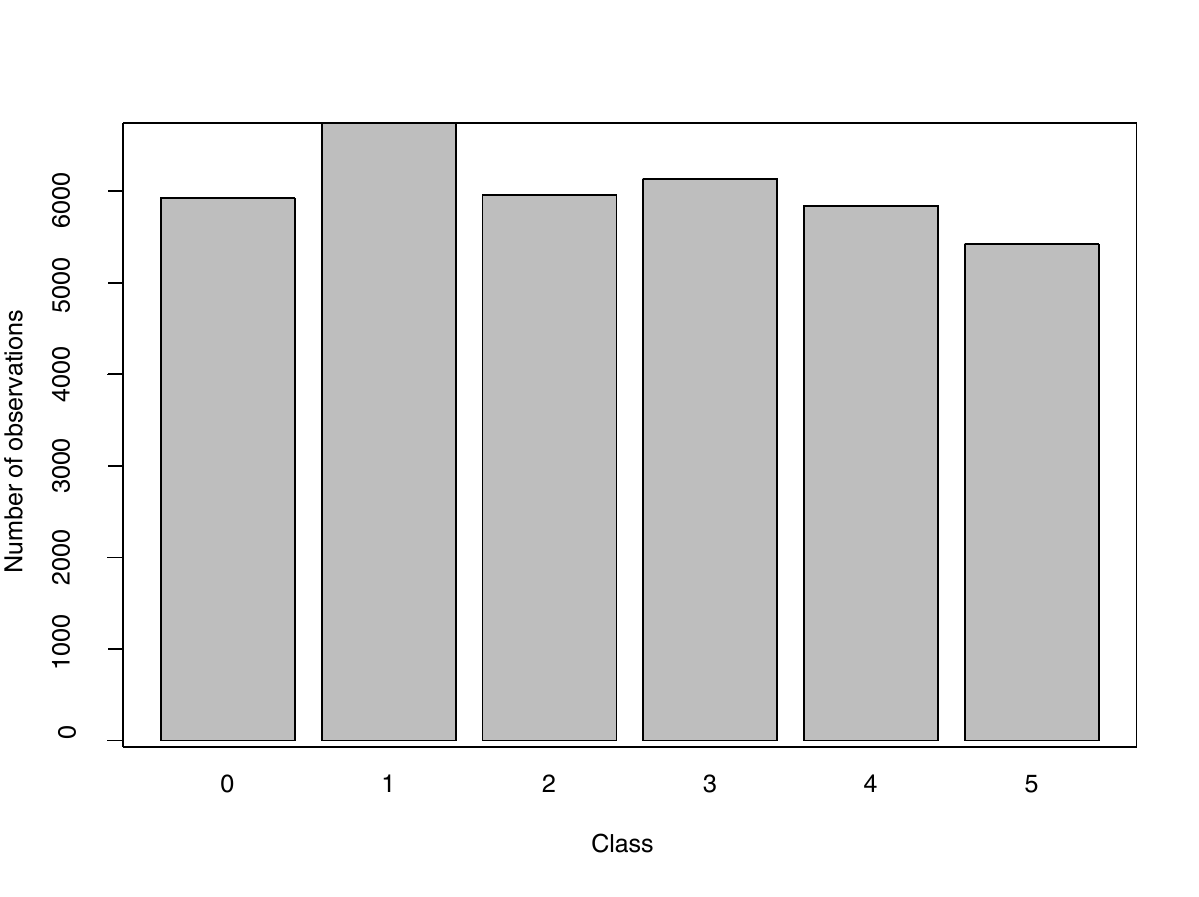}
		\caption{Class distribution in the MNIST training set after removing observations corresponding to digits $6$ through $9$.}
		\label{img:mnist-disttreino}
	\end{figure}

\begin{figure}[h]
		\centering
		\includegraphics[width=0.7\textwidth]{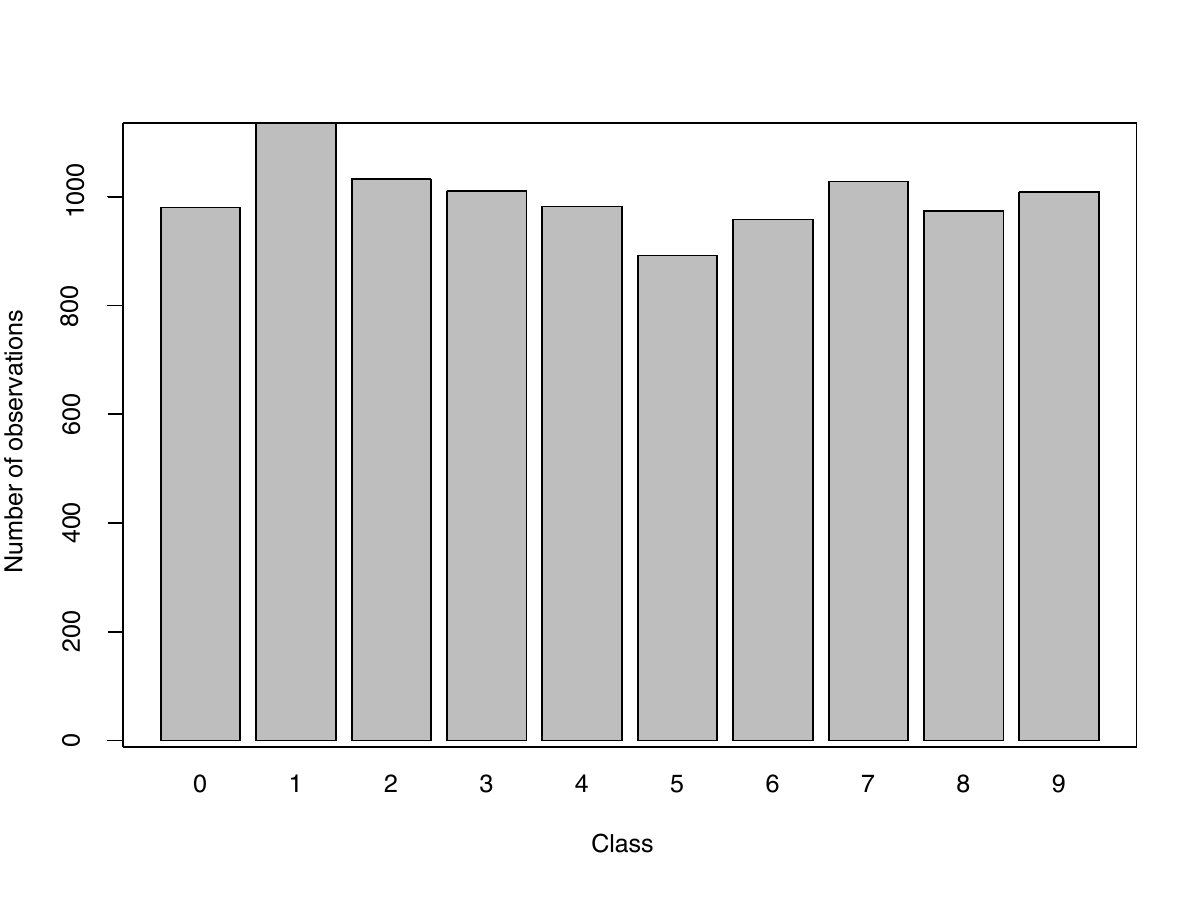}
		\caption{Class distribution in the MNIST test set.}
		\label{img:mnist-distteste}
	\end{figure}

	Figures~\ref{img:mnist-disttreino} and~\ref{img:mnist-distteste} show the class distributions in the MNIST dataset for the training and test sets, respectively. Note that we trained the algorithm using a subset of the original training set, containing $36{,}017$ observations of digits $0$ through $5$. The test set, on the other hand, includes $10{,}000$ observations of digits $0$ through $9$. Thus, digits $6$ through $9$ are considered outlier observations.
	
\begin{figure}[h]
		\centering
		\includegraphics[width=0.7\textwidth]{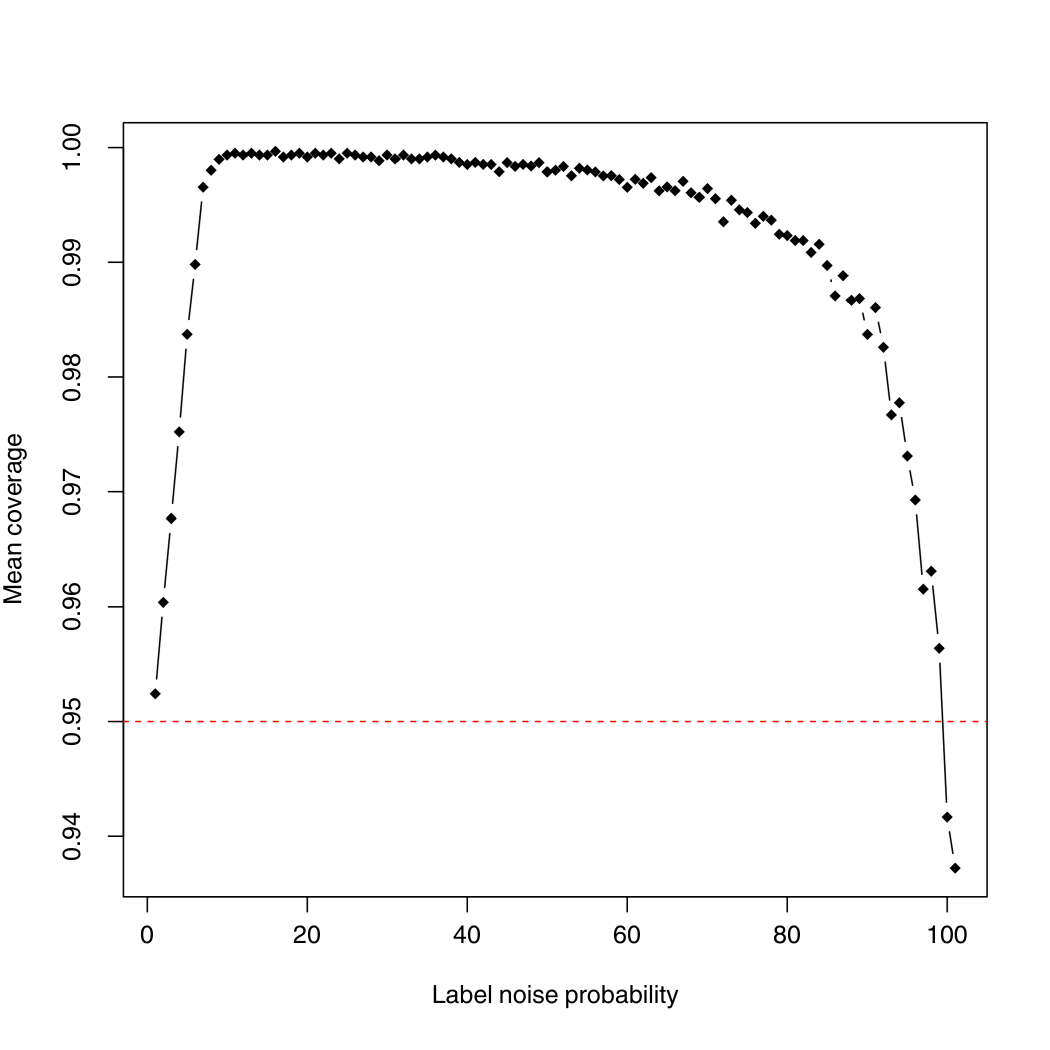}
		\caption{Evolution of the average coverage rate according to the noise level in the training set for the MNIST dataset.}
		\label{img:exmnist-rf-cobertura}
	\end{figure}

\begin{figure}[h]
		\centering
		\includegraphics[width=0.7\textwidth]{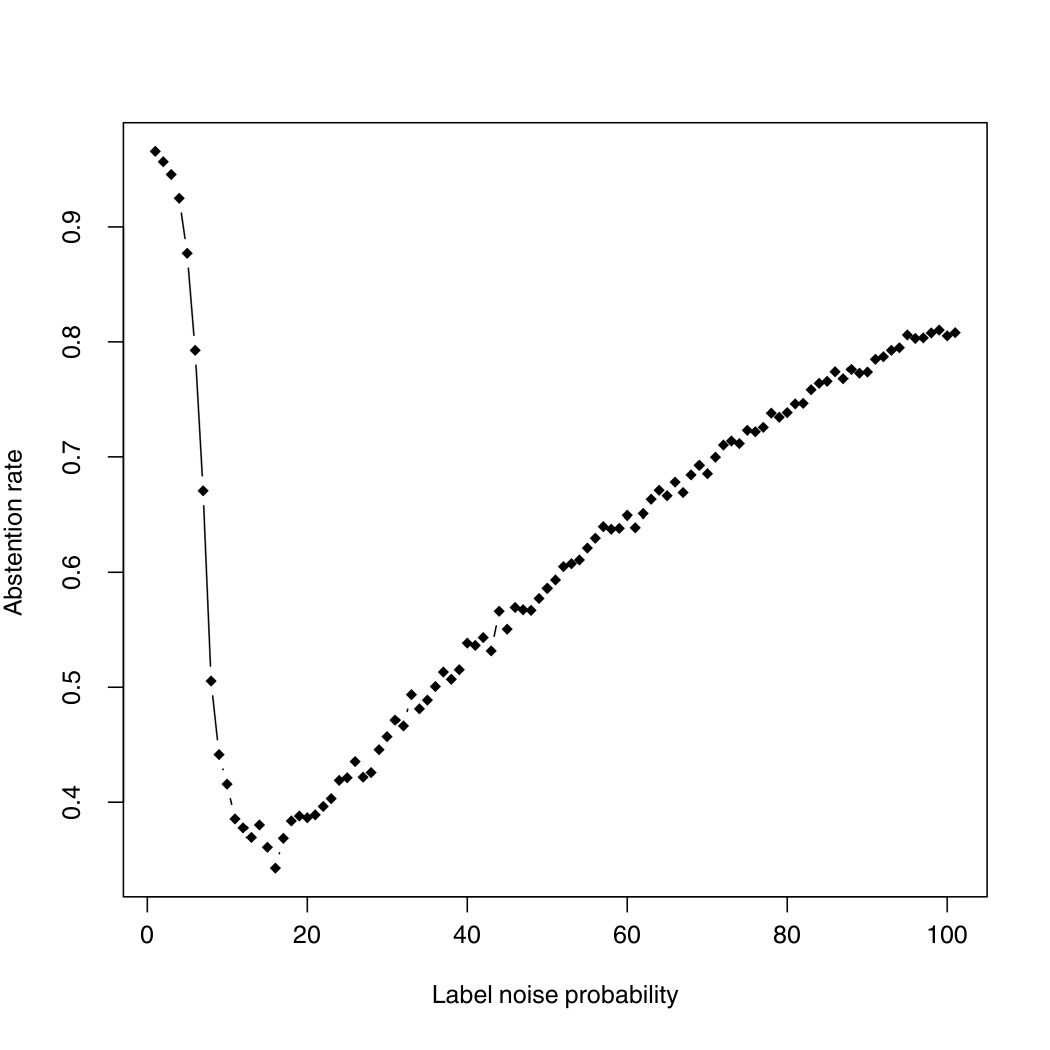}
		\caption{Evolution of the abstention rate for outlier observations according to the noise level in the training set for the MNIST dataset.}
		\label{img:exmnist-rf-abstencao}
	\end{figure}

	Figure~\ref{img:exmnist-rf-cobertura} shows the average class-wise coverage rates as a function of the noise level in the training labels. Figure~\ref{img:exmnist-rf-abstencao} presents the evolution of the abstention rate for outlier observations as the noise level in the training labels increases.
	
	\section{Discussion and Conclusion}
	\label{sec:conclusao}
	
	We can observe that the experiments conducted corroborate the findings of \citet{einbinder2022}, who argue that conformal prediction is robust to the inclusion of noise. For Example~\ref{ex1}, Figure~\ref{img:ex1-rf-cobertura} shows that the coverage rate remains stable for both classes up to approximately $40\%$ noise. For Examples~\ref{ex2} and~\ref{exmnist}, Figures~\ref{img:ex2-rf-cobertura} and~\ref{img:exmnist-rf-cobertura}, respectively, indicate that the average coverage almost never falls below the specified $95\%$ coverage guarantee. In fact, in the latter two examples, we can notice a slight increase in the average coverage even at low noise levels. However, this behavior might suggest that the BCOPS algorithm is including all or most of the classes in its prediction sets.
	
	Regarding the abstention rate for outlier observations, Figures~\ref{img:ex1-rf-abstencao}, \ref{img:ex2-rf-abstencao}, and \ref{img:exmnist-rf-abstencao} show that in all three experiments there is a significant drop in this metric for the first $10\%$ of noise, followed by a recovery as the noise level increases. This behavior indicates that the use of the BCOPS algorithm in scenarios where label errors may occur during training should be approached with caution.

	\section*{Acknowledgments}
	This research was carried out using HPC resources provided by the Superintendence of Information Technology at the University of São Paulo. The first author was supported by the São Paulo Research Foundation (FAPESP), grant no.~2022/11854-0. The second author was supported by the National Council for Scientific and Technological Development (CNPq), grant no.~304904/2020-1.

	\clearpage
	
	\bibliographystyle{my_apalike} 
	\bibliography{referencia.bib}

@article{guan2022,
	author =       {Guan, L. and Tibshirani, R.},
	title =        "Prediction and outlier detection in classification problems",
	journal =      "Journal of the Royal Statistical Society: Series B (Statistical Methodology)",
	volume =       "84",
	issue =        "2",
	pages =        "524--546",
	year =         "2022",
}

@article{lei2013,
	author =       {Lei, J. and Robins, J. and Wasserman, L.},
	title =        "Distribution-free prediction sets",
	journal =      "Journal of the American Statistical
	Association",
	volume =       "108",
	pages =        "278--287",
	year =         "2013",
}

@article{einbinder2022,
	author =       {Einbinder, B. and Bates, S. and Angelopoulos, A. and Gendler, A. and Romano, Y.},
	title =        "Conformal Prediction is Robust to Label Noise",
	journal =      "arXiv
	preprint arXiv:2209.14295",
	year =         "2022",
}

@article{lei2018,
	author = {Jing Lei and Max G’Sell and Alessandro Rinaldo and Ryan J. Tibshirani and Larry Wasserman},
	title = {Distribution-Free Predictive Inference for Regression},
	journal = {Journal of the American Statistical Association},
	volume = {113},
	pages = {1094-1111},
	year  = {2018},
}

@article{lei2015,
	author =       {Lei, J. and  Rinaldo, A. and Larry Wasserman},
	title =        "A Conformal Prediction Approach to Explore Functional Data",
	journal =      "Annals of Mathematics and Artificial Intelligence",
	year =         "2015",
	volume =       "74",
	pages =        "29--43",
}

@book{vovk2005,
	title={Algorithmic Learning in a Random World},
	author={Vovk, V. and Gammerman, A. and Shafer, G.},
	year={2005},
	publisher={Springer},
	address={New York}
}

@book{hastie2009,
	title={The Elements of Statistical Learning: Data Mining, Inference, and Prediction, Second Edition},
	author={Hastie, T. and Tibshirani, R. and Friedman, J.},
	lccn={2008941148},
	year={2009},
	publisher={Springer},
	address={New York}
}

@Manual{rlang,
	title = {R: A Language and Environment for Statistical Computing},
	author = {{R Core Team}},
	organization = {R Foundation for Statistical Computing},
	address = {Vienna, Austria},
	year = {2023},
	url = {https://www.R-project.org/},
}

@Article{ranger2017,
	title = {{ranger}: A Fast Implementation of Random Forests for High Dimensional Data in {C++} and {R}},
	author = {Marvin N. Wright and Andreas Ziegler},
	journal = {Journal of Statistical Software},
	year = {2017},
	volume = {77},
	number = {1},
	pages = {1--17},
}

@article{mnist,
	title={The mnist database of handwritten digit images for machine learning research},
	author={Deng, Li},
	journal={IEEE Signal Processing Magazine},
	volume={29},
	number={6},
	pages={141--142},
	year={2012},
	publisher={IEEE}
}

@Article{glmnet,
	title = {Regularization Paths for Generalized Linear Models via
	Coordinate Descent},
	author = {Jerome Friedman and Robert Tibshirani and Trevor Hastie},
	journal = {Journal of Statistical Software},
	year = {2010},
	volume = {33},
	number = {1},
	pages = {1--22},
}

@Article{gammerman98,
	title={Learning by transduction},
	author={Alex Gammerman and Volodya Vovk and Vladimir Vapnik},
	journal = {Proceedings of the Fourteenth Conference on Uncertainty in Artificial Intelligence},
	year={1998},
	pages = {148--155},
	publisher={Morgan Kaufmann Publishers Inc., San Francisco, CA}
}
	
\end{document}